\documentclass{article}
\usepackage[margin=1in]{geometry}
\usepackage[T1]{fontenc}
\usepackage{parskip}

\usepackage{microtype}
\usepackage{graphicx}
\usepackage{subfigure}
\usepackage{booktabs} 

\usepackage[hidelinks]{hyperref}

\usepackage{algorithm}
\usepackage{algorithmic}

\usepackage{amsmath}
\usepackage{amssymb}
\usepackage{mathtools}
\usepackage{amsthm}
\usepackage{bm}

\title{Efficiently predicting high resolution mass spectra\\with graph neural networks}
\author{
    Michael Murphy$^1$\footnote{The lead author carried out this work as an intern at Enveda Biosciences.}\\
    \texttt{murphy17@mit.edu}
    \and
    Stefanie Jegelka$^1$\\
    \texttt{stefje@mit.edu}
    \and
    Ernest Fraenkel$^1$\\
    \texttt{fraenkel@mit.edu}
    \and
    Tobias Kind$^2$\\
    \texttt{tobias.kind@envedabio.com}
    \and
    David Healey$^2$\\
    \texttt{david.healey@envedabio.com}
    \and
    Thomas Butler$^2$\\
    \texttt{tom.butler@envedabio.com}
}
\date{
    $^1$MIT\\
    $^2$Enveda Biosciences
     \\~\\
    January 26, 2023
}

\begin{document}

\maketitle






\begin{abstract}
\noindent Identifying a small molecule from its mass spectrum is the primary open problem in computational metabolomics.
This is typically cast as information retrieval: an unknown spectrum is matched against spectra predicted computationally from a large database of chemical structures.
However, current approaches to spectrum prediction model the output space in ways that force a tradeoff between capturing high resolution mass information and tractable learning.
We resolve this tradeoff by casting spectrum prediction as a mapping from an input molecular graph to a probability distribution over molecular formulas.
We discover that a large corpus of mass spectra can be closely approximated using a fixed vocabulary constituting only $2\%$ of all observed formulas.
This enables efficient spectrum prediction using an architecture similar to graph classification -- \textsc{GrAFF-MS} -- achieving significantly lower prediction error and orders-of-magnitude faster runtime than state-of-the-art methods.
\end{abstract}

\section{Introduction}

The identification of unknown small molecules in complex chemical mixtures is a primary challenge in many areas of chemical and biological science. 
The standard high-throughput approach to small molecule identification is \textit{tandem mass spectrometry} (MS/MS), with 
diverse applications including metabolomics \cite{dettmer2006metabolomics}, drug discovery \cite{atanasov2021natural}, clinical diagnostics \cite{evans2022predicting}, forensics \cite{brown2020forensics}, and environmental monitoring \cite{hernandez2012environmental}.

The key bottleneck in MS/MS is \textit{structural elucidation}: given a mass spectrum, we must determine the 2D structure of the molecule it represents. This problem is far from solved, and adversely impacts all areas of science that use MS/MS. Typically only $2{-}4\%$ of spectra are identified in untargeted metabolomics experiments \cite{da2015illuminating}, and a recent competition saw no more than 30\% accuracy \cite{casmi2022}.

Because MS/MS is a lossy measurement, and existing training sets are small, direct prediction of structures from spectra is particularly challenging. Therefore the most common approach is \textit{spectral library search}, which casts the problem as information retrieval \cite{stein2012libraries}: an observed spectrum is queried against a library of spectra with known structures. This provides an informative prior, and has the advantage of easy interpretability as the entire space of solutions is known. 
As there are relatively few $(10^4)$ small molecules with known experimental mass spectra, in spectral library search it is necessary to augment libraries with spectra predicted from large databases $(10^6-10^9)$ of molecular graphs.
This motivates the problem of \textit{spectrum prediction}.

Spectrum prediction is actively studied in metabolomics and quantum chemistry \cite{krettler2021map}, yet has received little attention from the machine learning community.
A major challenge in spectrum prediction is modelling of the output space: a mass spectrum is a variable-length set of real-valued ($m/z$, height) tuples, which is not straightforward to represent as an output of a deep learning model.
The $m/z$ coordinate (\textit{mass-to-charge ratio}) poses particular difficulty: it must be predicted with high precision, as a key strength of MS/MS is the ability to distinguish fractional $m/z$ differences on the order of $10^{-6}$ representative of different elemental composition.

Existing approaches to spectrum prediction force a tradeoff between capturing high-resolution $m/z$ information and tractability of the learning problem.
\textit{Mass-binning} methods \cite{wei2019neims,zhu2020preprint,young2021preprint} represent a spectrum as a fixed-length vector by discretizing the $m/z$ axis at regular intervals, discarding valuable information in favor of tractable learning.
\textit{Bond-breaking} methods \cite{wang2021cfmid,ruttkies2019metfrag} achieve perfect $m/z$ resolution, but do so through expensive combinatorial enumeration of substructures.

This work makes the following contributions:
\begin{itemize}
    \item We formulate spectrum prediction as a mapping from a molecular graph to a probability distribution over molecular formulas, allowing full resolution predictions without enumerating substructures;
    \item We discover most mass spectra can be effectively approximated with a small fixed vocabulary of molecular formulas, bypassing the tradeoff between $m/z$ resolution and tractable learning; and
    \item We implement an efficient graph neural network architecture, \textsc{GrAFF-MS}, that substantially outperforms state-of-the-art in both prediction error and runtime on two canonical mass spectrometry datasets.
\end{itemize}

\section{Background}

\subsection{Definitions}

We denote vectors $\bm{x}$ in bold lowercase and matrices $\bm{X}$ in bold uppercase. 

A \textit{molecular graph} $G = (V, E, \bm{a}, \bm{b})$ is a minimal description of the structure of a molecule: comprising an undirected graph with node set $V$, edge set $E \subset V \times V$, node labels $\bm{a} \in [118]^V$ indicating atom number, and edge labels $\bm{b} \in (\{1, 1.5, 2, 3\} \times \{-1, 0, 1\})^E$ indicating bond order and chirality. 

A \textit{molecular formula} $f$ (e.g. C$_{8}$H$_{10}N_{4}O_{2}$) describes a multiset of atoms, which we encode as a nonnegative integer vector of atom counts in $\mathcal{F}^* \doteq \mathbb{Z}_+^{118}$.
Formulas may be added and subtracted from one another, and inequalities between formulas are taken to hold elementwise.
We reserve the symbol $P$ to indicate a formula of a precursor ion.
The \textit{subformulas} of $P$ are the set $\mathcal{F}(P) \doteq \{f \in \mathcal{F}^* : f \le P\}$.

$\langle \mu, f \rangle \in \mathbb{R}_+$ is the \textit{theoretical mass} of a molecule with formula $f$, with units of daltons (Da): this is a linear combination of the monoisotopic masses of the elements of the periodic table, $\mu \in \mathbb{R}_+^{118}$, with multiplicities given by $f$.

A \textit{mass spectrum} $S$ is a variable-length set of \textit{peaks}, each of which is a ($m/z$, height) tuple $(m_i,y_i) \in \mathbb{R}_+^{2}$.
We use the notation $i \in S$ to index peaks in a spectrum.
We assume spectra are normalized, permitting us to treat them as probability distributions: $\sum_{i\in S} {y_i} = 1$.
A mass spectrum is implicitly  always accompanied by a precursor formula $P$.

We always assume charge $z\doteq 1$, as it is rare for small molecules to acquire more than a single charge.

\subsection{Tandem mass spectrometry}

\begin{figure}[ht!]
\begin{center}
\centerline{\includegraphics[width=0.67\columnwidth]{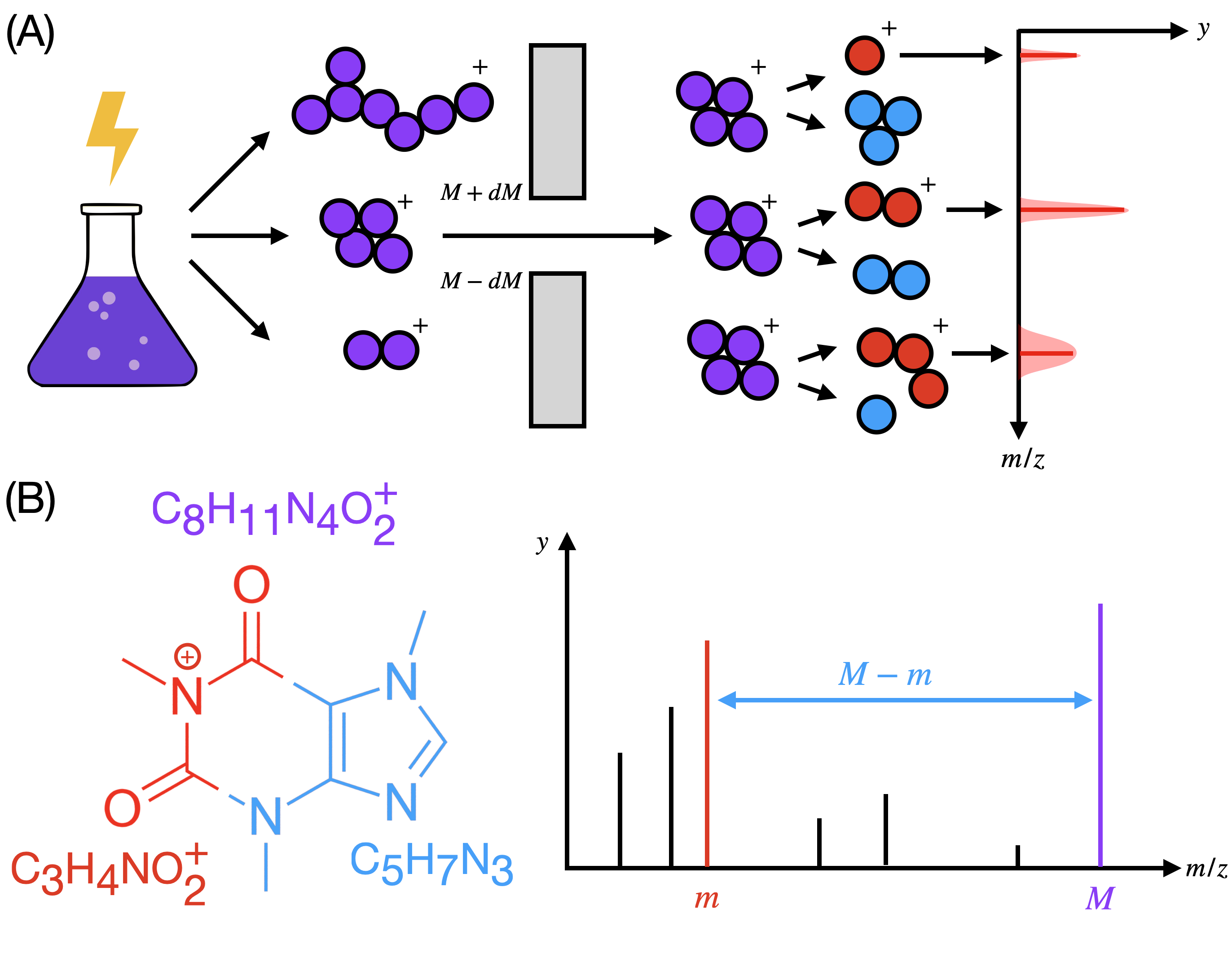}}
\caption{(A) The workflow of tandem mass spectrometry. A chemical mixture is ionized and filtered to isolate precursor ions of $m/z = M$; these are fragmented into product ions (red) and neutral losses (blue), and a detector yields a histogram of product ions indexed by $m/z$, with measurement error proportional to $m/z$. (B) An example fragmentation of a precursor ion with formula C$_8$H$_{11}$N$_4$O$_2$. Fragmentation breaks bonds, cutting the molecular graph into connected components. The component retaining the charge is the product ion; its complement is the neutral loss. The peak at $m/z = m$ represents the product ion; given the precursor formula, we can equally specify this peak by its formula C$_3$H$_4$NO$_2$, or the formula of its neutral loss C$_5$H$_7$N$_3$.}
\label{fig:mass-spec}
\end{center}
\end{figure}

A tandem mass spectrometer is a scientific instrument that generates high-throughput experimental signatures of the molecules present in a complex mixture.
It works by ionizing a chemical sample into a jet of electrically-charged gas.
This gas is electromagnetically filtered to select a population of \textit{precursor ions} of a specific mass-to-charge ratio ($m/z$) representing a unique molecular structure.
Each precursor ion is fragmented by collision with molecules of an inert gas.
If a collision occurs with sufficient energy, one or more bonds in the precursor will break, yielding a charged \textit{product ion} and one or more uncharged \textit{neutral loss} molecules.
The product ion is measured by a detector, which records its $m/z$ up to a small measurement error proportional to the $m/z$ times the instrument resolution $\epsilon$.
This process is repeated for large numbers of identical precursor ions, building up a histogram indexed by $m/z$.
Local maxima in this histogram are termed \textit{peaks}: ideally, each peak represents a unique product ion, with height reflecting its probability of formation.
This set of peaks constitutes the mass spectrum.
A typical mass spectrometry experiment acquires mass spectra for tens of thousands of distinct precursors in this manner, with $m/z$s measured at resolution on the order of $10^{-6}$. 
This process is depicted in Figure \ref{fig:mass-spec}A; we illustrate the relationship between precursor ion, product ion, and neutral loss in Figure \ref{fig:mass-spec}B.

\subsection{Structural elucidation}

A tandem mass spectrum of a molecule contains substantial information about its 2D structure.
Peak heights reflect the propensity of different bonds to break: this can reveal structural differences in compounds, even if they have the same molecular formula. 
Furthermore, the $O(10^{-6})$ resolution of modern MS/MS can detect small characteristic deviations from integrality in the masses of individual chemical elements that arise from nuclear physics, allowing detailed information about the molecular formula to be inferred from the spectrum with high accuracy (Sec \ref{sec:massdecomp}).

The task of inferring a 2D structure from a mass spectrum, \textit{structural elucidation}, is the major open problem in computational metabolomics, dating back to the 1960s in one of the earliest examples of an expert system \cite{lindsay1993}. 
Yet structural elucidation remains far from solved: despite half a century of research, algorithmic approaches perform only marginally better than manual annotations by expert chemists, with no method submitted to the 2022 Critical Analysis of Small Molecule Identification (CASMI) challenge exceeding $30\%$ accuracy \cite{casmi2022}.

The difficulty of structural elucidation has broad scientific implications: while a modern metabolomics experiment routinely detects tens of thousands of spectra, usually only a few percent of these are confidently annotated with a structure \cite{da2015illuminating}.
Spectral library search, described in the introduction, is the approach to structural elucidation preferred in practice by most biologists and chemists \cite{stein2012libraries}, and is a standard component of existing metabolomics workflows. The availability of high-quality predicted spectra across large chemical structure databases would therefore greatly increase compound identification rates in real experimental settings. 



\subsection{Mass decomposition} \label{sec:massdecomp}

Modern mass spectrometry achieves sufficiently high resolution to detect small deviations in $m/z$ from integrality that are characteristic of different chemical elements.
This property is a key strength of the technology, because it permits annotating peaks with formulas through \textit{mass decomposition} \cite{duhrkop2013decomposition}.
Given a product ion of $m/z = m$, a precursor formula $P$, and an instrument resolution $\epsilon$, \textit{product mass decomposition} yields a set $\mathcal{F}(P,m,\epsilon)$ of chemically plausible subformulas of $P$ whose theoretical masses lie within measurement error $\epsilon m$ of $m$.
This can be cast as an integer program, in which all solutions of the following minimization with cost $\le \epsilon m$ are enumerated:
\begin{align}
    \min_{f\in\mathcal{F}^*} &\  \vert \langle\mu,f\rangle - m \vert \\
    \text{s.t.} &\  f \le P, f \in \Omega
\end{align}
where $\Omega$ describes a general set of constraints that exclude unrealistic molecular formulas \cite{kind2007sgr}.
In practice, with modern instruments $\epsilon$ is sufficiently small for there to typically be only one or a few solutions to this optimization.
This allows us to later rely on product mass decomposition as a black-box to generate useful formula annotations at training time.

\section{Related work}

\textit{Bond-breaking}, used in \cite{wang2021cfmid,ruttkies2019metfrag,cao2021moldiscovery}, solves the problem of representing the output space by enumerating the 2D structures of all probable product ions.
These are taken to be connected subgraphs of the precursor, generated by sequences of edge removals.
Each product ion structure is scored for its probability of formation, and a spectrum is generated by associating this probability with each structure's theoretical $m/z$.
Bond-breaking therefore achieves perfect $m/z$ resolution, but suffers from two major weaknesses: first, enumerating substructures scales poorly with molecule size, and is not conducive to massively-parallel implementation on a GPU.
We found a state-of-the-art method \cite{wang2021cfmid} takes ${\sim}5$s on average to predict a single mass spectrum, which precludes training on the largest available datasets: using the same settings as its authors, training \cite{wang2021cfmid} on $\sim$300k spectra in \textsc{NIST-20} would take an estimated \textit{three months} on a 64-core machine. 
It also poses serious limitations at test time, as inference with a large-scale structure database like ChEMBL \cite{gaulton2016chembl} requires predicting millions of spectra.
The other weakness of bond-breaking arises from a restrictive modelling assumption: \textit{rearrangement reactions} \cite{mclafferty1959rearrangement} frequently yield product ions that are not reachable from the precursor by sequences of edge removals.\footnote{While engineered rules are used in bond-breaking to account for certain well-studied rearrangements, we found the state-of-the-art method CFM-ID still fails to assign a formula annotation within 10ppm to $42\%$ of monoisotopic peaks in the \textsc{NIST-20} dataset.}

\textit{Mass-binning} is used for spectrum prediction by \cite{wei2019neims}, and subsequently employed in recent preprints \cite{zhu2020preprint,young2021preprint}.
This approach represents a mass spectrum as a fixed-length vector via discretization: the \textit{m/z} axis is partitioned into narrow regularly-spaced bins, and each bin is assigned the sum of the intensities of all peaks falling within its endpoints.
Spectrum prediction then becomes a vector-valued regression problem.
Mass-binning is conducive to GPU implementation and scales better than bond-breaking, but because a target space with millions of mass bins is too large, realistic bin counts lose essential high resolution information about the molecular formulas of the peaks: discarding a key strength of MS/MS analysis in favor of a tractable learning problem. 
Such models are also susceptible to edge effects, where $m/z$ measurement error of the instrument can cause peaks of the same product ion to cross bin boundaries from spectrum to spectrum.

Other approaches include molecular dynamics simulation \cite{koopman2021qcxms}, which has extremely high computational costs; and NLP-inspired models for peptides \cite{zhou2017pdeep,gessulat2019prosit}, which are effective but inapplicable to other types of molecules.
%
%
%
%

\section{\textsc{GrAFF-MS}}

Our approach constitutes three major components: first, we represent the output space of spectrum prediction as a space of probability distributions over molecular formulas.
We then introduce a constant-sized approximation of this output space using a fixed vocabulary of formulas, which we can generate from our training data; we later show this introduces only minor approximation cost, as most formulas occur with low probability.
Finally, we derive a loss function that takes into account data-specific ambiguities introduced by our model of the output space.
These components together allow us to efficiently predict spectra using a standard graph neural network architecture.  
We call our approach \textsc{GrAFF-MS}: (Gr)aph neural network for (A)pproximation via (F)ixed (F)ormulas of (M)ass (S)pectra.

\subsection{Modelling spectra as probability distributions over molecular subformulas of the precursor}

Our aim is to predict a mass spectrum from a molecular graph.
To do so, we must determine how to best represent the output space: a spectrum comprises a variable-length set of peaks located at continuous $m/z$ positions, whose heights sum to one.
We notice that peaks are not located arbitrarily: the set of $m/z$s is structured, as the $m/z$ of a peak is determined (up to measurement error) by the molecular formula of its corresponding product ion.
This formula is sufficient to determine the $m/z$; in particular, we do not need to know the product ion's full 2D structure.
We therefore model a mass spectrum as a probability distribution over \textit{molecular subformulas} $\mathcal{F}(P)$ of the precursor $P$:
\begin{align}
    S &= \{(m_f, y_f) : f \in \mathcal{F}(P)\}
\end{align}
where $m_f \doteq \langle \mu, f \rangle$ is the theoretical mass of formula $f$.

This is more efficient in principle than bond-breaking, which models a spectrum as a distribution over at worst exponentially many substructures of the precursor.
But the number of subformulas is only polynomial in the coefficients of the precursor formula -- and the majority of subformulas can be ruled out \textit{a priori} as chemically infeasible \cite{kind2007sgr}. 
It is also less restrictive than bond-breaking, which relies on hand-engineered rules to capture rearrangement reactions: enumerating subformulas is guaranteed to generate all possible peaks, irrespective of whether the structure of their product ion is reachable by edge removals or not. 
Yet our approach preserves the core advantage of bond-breaking over mass-binning: predicting a height for each subformula yields spectra with perfect $m/z$ resolution.

\subsection{Fixed-vocabulary approximation of formula space}

In practice, enumeration of subformulas is still a costly operation for larger molecules.
One way to avoid this would be to sequentially decode formulas of nonzero probability one at a time: we opt not to do so, as this requires a more complex, data-hungrier model, and necessitates a linear ordering of formulas, for which there is not an obvious correct choice. Instead, we exploit a property of small molecule mass spectra that we discovered in this work, and illustrate in Figure \ref{fig:vocab} of the results: almost all of the signal in small molecule mass spectra lies in peaks that can be explained by a relatively small number $({\sim}2\%)$ of product ion and neutral loss formulas that frequently recur across spectra.

Inspired by this finding, we approximate $\mathcal{F}(P)$ via the union $\hat{\mathcal{F}}(P) = \hat{\mathcal{P}} \cup (P - \hat{\mathcal{L}})$ of a fixed set of frequent product ion formulas $\hat{\mathcal{P}}$, and a variable set of `precursor-specific' formulas $P - \hat{\mathcal{L}}$ obtained by subtracting a fixed set of frequent neutral loss formulas $\hat{\mathcal{L}}$ from the precursor $P$.
This greatly simplifies the spectrum prediction problem: we now only need to predict a probability for each of the formulas in $\hat{\mathcal{P}}$ and $\hat{\mathcal{L}}$, which we can accomplish with time \textit{constant} in the size of the precursor.

Stated explicitly, we approximate the spectrum as:
\begin{align}
    S \approx \{(m_f, y_f) : f \in \hat{\mathcal{F}}(P) \}
\end{align}
where a height of zero is implicitly assigned to any formula not in $\hat{\mathcal{F}}(P)$.

The fact that we can equally represent a product ion by either its own formula or a neutral loss formula relative to its precursor ion is crucial to generalization.
If we only included frequent product ion formulas, we would explain peaks of low mass well, which typically correspond to small charged functional groups.
But as formula space becomes larger with increasing mass, it becomes increasingly unlikely that every significant peak of higher mass in an unseen compound will be explained.
However such peaks do not represent arbitrary subformulas of the precursor: they tend to arise from losses of small uncharged functional groups and combinations thereof, which we capture by including frequent neutral losses.

Our algorithm to generate $\hat{\mathcal{P}}$ and $\hat{\mathcal{L}}$ involves listing all product ion and neutral loss formulas yielded by mass decomposition of the training set, and ranking them by the sum of the heights of all peaks to which each formula is assigned; we select the top $K$ highest ranked among either type.
Pseudocode is provided in appendix \ref{selection}.

\subsection{Peak-marginal cross entropy}

To train our approach, we must rely on formula annotations generated by mass decomposition.
Because mass spectrometers have limited resolution, it is often the case that more than one valid subformula has a mass within measurement error of a peak.
These are considered equiprobable \textit{a priori}, and need not be mutually exclusive: it is possible for a compound to contain two distinct substructures with $m/z$ difference smaller than the measurement error.
As we cannot pick a single formula in such cases, we approximate the full cross entropy by marginalizing over compatible formulas: this yields the \textit{peak-marginal cross entropy}, which we minimize w.r.t. the parameters of a neural network $\hat{\bm{y}}(\cdot;\theta)$:
\begin{align} \label{eq:pmce}
    \min_{\theta}\ - \sum_{n=1}^N\sum_{i \in S_n} y_i^n \log \sum_{f\in \hat{\mathcal{F}}^n_i} \hat{y}_f(G_n;\theta)
\end{align}
using the shorthand $\hat{\mathcal{F}}^n_i \doteq \hat{\mathcal{F}}(P_n) \cap \mathcal{F}(P_n,m^n_i,\epsilon)$ to indicate the intersection of our approximated vocabulary with the formula annotations generated by mass decomposition.
We provide a derivation from first principles in appendix \ref{derivation}.

In this formulation, given a molecular graph $G$ of a precursor with formula $P$, our model predicts a probability $\hat{y}_f$ for every formula $f$ in the fixed vocabulary.
This produces a spectrum $\hat{S} = \{(m_f,\hat{y}_f) : f \in \hat{\mathcal{F}}(P)\}$.
These per-formula probabilities are summed within each observed peak across its compatible formulas to yield a predicted peak height, and the cross-entropy between the observed and predicted peak heights across the entire spectrum is minimized. 

\subsection{Model architecture}

Formulating spectrum prediction as graph classification permits a fairly typical GNN architecture. 
\textsc{GrAFF-MS} uses a graph isomorphism network with edge and graph Laplacian features \cite{xu2019gin,hu2020pretraining,lim2022signnet}: this encodes the molecular graph into a dense vector representation, which is then conditioned on mass-spectral covariates and passed through a feed-forward network that decodes a logit for each formula in the vocabulary.

We start with the graph of the 2D structure $G = (V, E)$, to which we add a virtual node \cite{gilmer2017quantum} and four classes of features: node features $\bm{a}_i \in \mathbb{R}^{d_{atom}}$, edge features $\bm{b}_{ij} \in \mathbb{R}^{d_{bond}}$, covariate features $\bm{c} \in \mathbb{R}^{d_{cov}}$, and the top eigenvectors and eigenvalues of the graph Laplacian $\bm{v}_i \in \mathbb{R}^{d_{eig}}, \bm{\lambda} \in \mathbb{R}^{d_{eig}}$.
We use the canonical atom and bond featurizers from DGL-LifeSci \cite{li2021dgllife} to generate $\bm{a}$ and $\bm{b}$.
Since a mass spectrum is not fully determined by the molecular graph, $\bm{c}$ includes a number of necessary experimental parameters: normalized collision energy, precursor ion type, instrument model, and presence of isotopic peaks.
Further details are provided in Table \ref{tab:covs} of the appendix; there we also provide hyperparameter settings. 

We first embed the node, edge, and covariate features into $\mathbb{R}^{d_{enc}}$, reusing the following \texttt{MLP} block:
\begin{align*}
    \texttt{MLP}(\cdot) = \texttt{LayerNorm}(\texttt{Dropout}(\texttt{SiLU}(\texttt{Linear}(\cdot))))
\end{align*}
and transform the Laplacian features into node positional encodings in $ \mathbb{R}^{d_{enc}}$ using a SignNet \cite{lim2022signnet} with $\phi$ and $\rho$ both implemented as 2-layer stacked \texttt{MLP} blocks:
\begin{align}
    \bm{x}_i^{atom} &= \texttt{MLP}_{atom}(\bm{a}_i) \\
    \bm{x}_i^{eig} &= \texttt{SignNet}(\bm{v}_i, \bm{\lambda}) \\
    \bm{x}_{ij}^{bond} &= \texttt{MLP}_{bond}(\bm{b}_{ij}) \\
    \bm{x}^{cov} &= \texttt{MLP}_{cov}(\bm{c})
\end{align}
taking $i \in V$ and $(i,j) \in E$.
We sum the embedded atom features and node positional encodings, and pass these along with embedded bond features into a stack of alternating $L$ message-passing layers to update the node representations, and $L$ \texttt{MLP} layers to update the edge representations.
\begin{align}
    \bm{x}_i^{(0)} &= \bm{x}_i^{atom} + \bm{x}_i^{eig} \\
    \bm{e}_{ij}^{(0)} &= \bm{x}_{ij}^{bond} \\
    \bm{X}^{(l+1)} &= \bm{X}^{(l)} + \texttt{GINEConv}^{(l)}(G,\bm{X}^{(l)}, \bm{E}^{(l)}) \\
    \bm{e}_{ij}^{(l+1)} &= \bm{e}_{ij}^{(l)} + \texttt{MLP}^{(l)}_{edge}(\bm{e}^{(l)}_{ij} \Vert \bm{x}^{(l+1)}_{i} \Vert \bm{x}^{(l+1)}_{j})
\end{align}
where $\Vert$ denotes concatenation. The message-passing layer uses the \texttt{GINEConv} operation implemented in \cite{fey2019pyg}: for its internal feed-forward network, we use two stacked \texttt{MLP} blocks with GraphNorm \cite{cai2021graphnorm} in place of layer normalization.
We similarly replace layer normalization with GraphNorm in the $\texttt{MLP}_{edge}$ blocks.
Both node and edge updates use residual connections, which we found greatly accelerate training.

We generate a dense representation of the molecule by attention pooling over nodes \cite{er2016pooling}, to which we add the embedded covariate features.
This is decoded by a feed-forward network into a spectrum representation $\bm{x}^{spec} \in \mathbb{R}^{d_{dec}}$.
\begin{align}
    a_i &= \texttt{Softmax}_{i\in V}(\texttt{Linear}(\bm{x}_i)) \\
    \bm{x}^{mol} &=  \sum_{i\in V} a_i \bm{x}^{(L)}_i \\
    \bm{x}^{spec} &= \texttt{MLP}_{spec}(\bm{x}^{mol} + \bm{x}^{cov})
\end{align}
where $\texttt{MLP}_{spec}$ is a stack of $L'$ \texttt{MLP} blocks with residual connections. 
In principle we may now project this representation via a linear layer $(\bm{w}_k,b_k)$ into a logit $z_k$ for each of the $K$ product ion or neutral loss formulas in the vocabulary. 

\subsection{Domain-specific modifications}

We must now introduce a number of corrections motivated by domain knowledge to produce realistic mass spectra. 

Following fragmentation, certain product ions can bind ambient water or nitrogen molecules as \textit{adducts}, shifting the mass of the fragment.
This occurrence is annotated in our training set.
For each formula in our vocabulary, we therefore predict a logit for three different adduct states of the fragment, indexed by $\alpha$: the original product ion $f$, $f$ $+$ H$_2$O, and $f$ $+$ N$_2$.
In effect this triples our vocabulary size.

Depending upon instrument parameters, tandem mass spectra can display small peaks arising from higher isotopic states of the precursor ion, at integral $m/z$ shifts relative to the monoisotopic peak.
Isotopic state is independent of fragmentation: so rather than expanding our vocabulary again, we apply to all predictions a shared offset for each isotopic state $\beta \in \{0, 1, 2\}$, which we parameterize on $\bm{x}^{spec}$.

As our vocabulary includes both product ions and neutral losses, we must deal with occasional \textit{double-counting}: depending on the precursor $P$, there are cases where the same subformula $f$ will be predicted both as a product ion ($f \in \hat{\mathcal{P}}$) and a neutral loss ($P - f \in \hat{\mathcal{L}}$).
In such cases -- denoted by the set $\hat{\mathcal{D}}(P)$ -- we subtract a $\log 2$ correction factor from both logits: this way the innermost summation in Equation \ref{eq:pmce} takes the mean of their contributions instead of their sum. 

Applying these corrections and softmaxing yields the final heights of the predicted mass spectrum $\hat{\bm{y}}$:
\begin{align}
    z_{k}^{\alpha\beta} &= \bm{w}_{k}^{\alpha}\bm{x}^{spec} + \bm{w}^\beta\bm{x}^{spec} +  b_{k}^{\alpha} \\ &- \mathbb{I}[k \in \hat{\mathcal{D}}(P)] \log 2 \\
    \hat{y}_{k}^{\alpha\beta} &= 
    \texttt{Softmax}_{k,\alpha,\beta}(z_{k}^{\alpha\beta})
\end{align}
\section{Experiments}

\subsection{Datasets}

\subsubsection{NIST-20}

We train our model on the \textsc{NIST-20} tandem MS spectral library \cite{nist20}.
This is the largest dataset of high resolution mass spectra of small molecules, curated by expert chemists, and is commercially available for a modest fee.\footnote{Open data is not the norm in small molecule mass spectrometry, as large-scale annotation requires substantial time commitment from teams of highly-trained human experts.
As a result, no public-domain dataset exists of comparable scale and quality to \textsc{NIST-20}. 
However, \textsc{NIST-20} and its predecessors are well-established in academic mass spectrometry, and have been used in previous machine learning publications \cite{wei2019neims,duhrkop2022kernel}.}
For each measured compound, \textsc{NIST-20} provides typically several spectra acquired across a range of collision energies. 
Each spectrum is represented as a list of ($m/z$, intensity, annotation) peak tuples, in addition to metadata describing instrumental parameters and compound identity.
The annotation field includes a list of formula hypotheses per peak that were computed via mass decomposition.

We restrict \textsc{NIST-20} to HCD spectra with $[M+H]^+$ or $[M-H]^-$ precursor ions.
We exclude structures that are annotated as glycans or peptides or exceed 1000Da in mass (as these are not typically considered small molecules), or have atoms other than \{C, H, N, O, P, S, F, Cl, Br, I\}.

We use an $85/5/10$ structure-disjoint train/validation/test split, which we generate by grouping spectra according to the connectivity substring of their InChiKey \cite{heller2015inchi} and assigning spectra to splits an entire group at a time.
As CFM-ID only predicts monoisotopic spectra at qualitative energy levels \{low, medium, high\}, we restrict the test set to spectra with corresponding energies $\{20, 35, 50\}$ in which no peaks were annotated as higher isotopes.
This yields 306,135 (19,871) training, 17,987 (1,167) validation, and 4,548 (1,637) test spectra (structures).

\subsubsection{CASMI-16}

It is well known that uniform train-test splitting can overestimate generalization in molecular machine learning \cite{wu2018moleculenet}.
To address this issue, we employ an independent test set: the spectra of the 2016 CASMI challenge \cite{schymanski2017casmi}.
This is a small public-domain mass spectrometry dataset, constructed by domain experts specifically for testing algorithms, and comprises structures selected as representative of those encountered `in the wild' when performing mass spectrometry of small molecules. 

We use $[M+H]^+$ and $[M-H]^-$ spectra from the combined `Training' and `Challenge' splits from Categories 2 and 3 of the challenge.
We exclude any structures from \textsc{CASMI-16} with an InChiKey connectivity match to any in \textsc{NIST-20}, yielding $156$ spectra of $141$ structures.
The collision energy stepping protocol used in \textsc{CASMI-16} is simulated by predicting a spectrum at each of $\{20, 35, 50\}$ normalized collision energy and returning their mean.

\subsection{Baselines}

\subsubsection{CFM-ID}

CFM-ID \cite{wang2021cfmid} is a bond-breaking method, viewed by the mass spectrometry community as the state-of-the-art in spectrum prediction \cite{krettler2021map}.
We found CFM-ID prohibitively expensive to train on \textsc{NIST-20} (one parallelized EM iterate on a subset of $\sim$60k spectra took 10 hours on a 64-core machine) so we use trained weights provided by its authors, learned from 18,282 spectra in the commercial \textsc{METLIN} dataset \cite{guijas2018metlin}.
Domain experts consider spectra acquired under \textsc{METLIN}'s conditions interchangeable with those of \textsc{NIST-20} \cite{deleoz2018fragmentation} so it is reasonable to evaluate their model on our data.

\subsubsection{NEIMS}

NEIMS \cite{wei2019neims} is a feed-forward network that inputs a precomputed molecular fingerprint and outputs a mass-binned spectrum, which is postprocessed using a domain-specific gating operation.
We retrained NEIMS on \textsc{NIST-20}, which necessitated two modifications: (1) we concatenate a vector of covariates to the fingerprint vector, without which \textsc{NIST-20} spectra are not fully determined; and (2) we bin at 0.1Da intervals instead of 1Da intervals, to account for finer instrument resolution in \textsc{NIST-20}.
We otherwise use the same hyperparameter settings as the original paper, and early-stop on validation loss.

\subsection{Evaluation metrics}

We quantify predictive performance using \textit{mass-spectral cosine similarity}, which compares spectra subject to $m/z$ measurement error \cite{bittremieux2022cosine} by matching pairs of peaks.
For two spectra $S$ and $\hat{S}$, mass-spectral cosine similarity $C_{S,\hat{S}}$ is the value of the following linear sum assignment problem:
\begin{align}
    C_{S,\hat{S}} \doteq \max_{x_{ij} \in \{0,1\}} & \sum_{\substack{i\in S,\ j\in \hat{S} : \\ \vert m_i - \hat{m}_j \vert \le \tau}}x_{ij}\frac{y_i}{\Vert y \Vert_2}\frac{\hat{y}_j}{\Vert \hat{y} \Vert_2} \\
    \text{s.t. } & \textstyle\sum_{i\in S} x_{ij} \le 1 \\
    & \textstyle\sum_{j\in \hat{S}} x_{ij} \le 1
\end{align}
We use the \texttt{CosineHungarian} implementation in the \texttt{matchms} Python package, with tolerance $\tau = 0.05$.
We report mean cosine similarity across spectra, as well as the fraction of spectra scoring $> 0.7$, which is commonly employed in spectral library search as a heuristic cutoff for a positive match \cite{wohlgemuth2016splash}.

We also compare runtime of \textsc{GrAFF-MS} against the bond-breaking method CFM-ID.
For fair comparison, we time a forward pass for each structure in the \textsc{NIST-20} test split using only the CPU, without any batching.
We include time spent in preprocessing: our input is a SMILES string and experimental covariates, and our output is a spectrum.
As collision energy affects the number of peaks that CFM-ID generates, we predict spectra at low, medium, and high energies and use the average runtime of the three.



\section{Results}

\subsection{A fixed vocabulary of product ions and neutral losses closely approximates most mass spectra}

\begin{figure}[ht!]
\begin{center}
\centerline{\includegraphics[width=0.5\columnwidth]{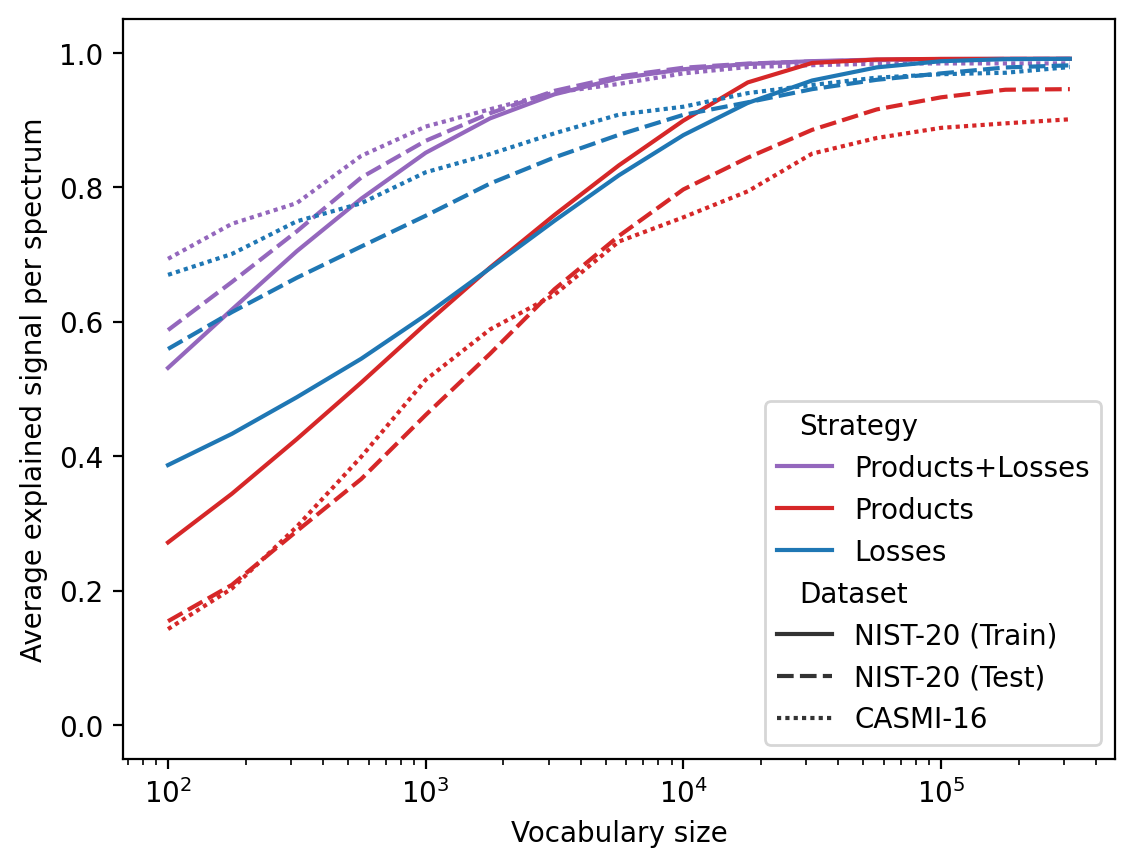}}
\caption{Generalization of different heuristics for fixed-size vocabulary selection. For a given vocabulary size on the x-axis, the y-axis indicates the sum of all explained peaks' heights within a given spectrum, averaged over all spectra.}
\label{fig:vocab}
\end{center}
\end{figure}

Figure \ref{fig:vocab} demonstrates the fraction of ion counts explained in the average mass spectrum as the training vocabulary size is varied.
This shows most signal lies within peaks explainable by a relatively small number of product ion and neutral loss formulas.
In particular, the vocabulary we use (of size $K = 10^4$) is sufficient to explain $98\%$ of ion counts in the \textsc{NIST-20} training split, which comprises 193,577 unique product ion formulas and 348,692 unique neutral loss formulas.
We observe that a fixed vocabulary generalizes beyond \textsc{NIST-20} to a separate dataset \textsc{CASMI-16}, indicating `formula sparsity' is a general property of small molecule mass spectra and not limited to our particular training set.
We also compare alternative strategies of picking only the top products or top losses: using both types of formulas explains more signal for a given $K$ than either alone.

\subsection{\textsc{GrAFF-MS} outperforms bond-breaking and mass-binning on standard MS/MS datasets}

Table \ref{tab:cosine} shows \textsc{GrAFF-MS} produces spectra with greater cosine similarity to ground-truth than either baseline.
More of our spectra also meet the $C_{S\hat{S}} > 0.7$ threshold for useful predictions.
These results hold both for the \textsc{NIST-20} test split and the independent test set \textsc{CASMI-16}.
We see all methods perform better on \textsc{CASMI-16} than \textsc{NIST-20}: this is because \textsc{NIST-20} includes a minority of substantially larger molecules (max weight $986$Da) than \textsc{CASMI-16} (max weight $539$Da), with which all three methods struggle.

\begin{table}[ht]
\caption{Mean cosine similarity $\mathbb{E}[C]$ and fraction of useful predictions $\mathbb{P}(C > 0.7)$ on the \textsc{NIST-20} test split and \textsc{CASMI-16}. $95\%$ confidence intervals are computed via nonparametric bootstrap.}
\label{tab:cosine}
\begin{center}
\begin{small}
\begin{sc}
\begin{tabular}{lllll}
\toprule
&
  \multicolumn{2}{c}{\begin{tabular}[c]{@{}c@{}}NIST-20 Test\\ (N = 4548)\end{tabular}} &
  \multicolumn{2}{c}{\begin{tabular}[c]{@{}c@{}}CASMI-16\\ (N = 156)\end{tabular}} \\
 Method &
  \multicolumn{1}{c}{$\mathbb{E}[C]$} &
  \multicolumn{1}{c}{$\mathbb{P}(C{>}0.7)$} &
  \multicolumn{1}{c}{$\mathbb{E}[C]$} &
  \multicolumn{1}{c}{$\mathbb{P}(C{>}0.7)$} \\
\midrule
CFM-ID & .52$\pm$.01                & .35$\pm$.02               & .75$\pm$.05 & .70$\pm$.07 \\
NEIMS  & .60$\pm$.01                & .50$\pm$.01               & .63$\pm$.05                      & .54$\pm$.08                       \\
\textsc{GrAFF-MS} & \textbf{.70}$\pm$.01                & \textbf{.62}$\pm$.02              & \textbf{.79}$\pm$.05 & \textbf{.76}$\pm$.07 \\
\bottomrule
\end{tabular}
\end{sc}
\end{small}
\end{center}
\end{table}

\subsection{Representing peaks as formulas scales better with molecular weight than bond-breaking}

\begin{figure}[ht]
\begin{center}
\centerline{\includegraphics[width=0.5\columnwidth]{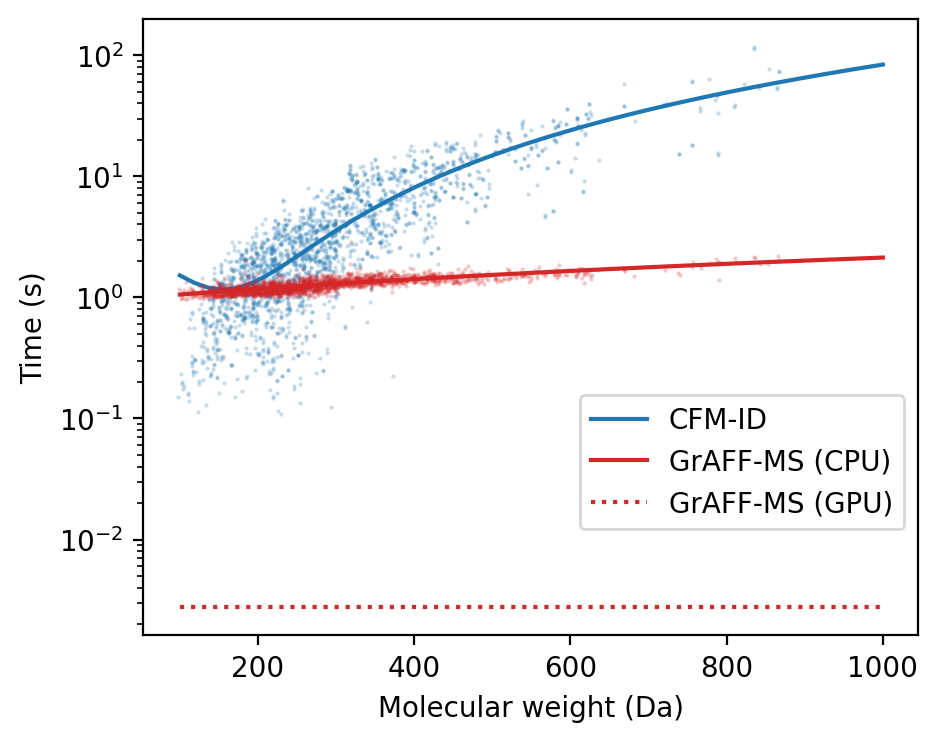}}
\caption{Empirical time complexity on \textsc{NIST-20} structures with respect to molecular weight. Each dot is a structure. Solid lines are quadratic (blue) and linear (red) fits; dotted line indicates an average over all spectra computed using shuffled minibatches.}
\label{fig:time}
\end{center}
\end{figure}

Figure \ref{fig:time} shows our approach to modelling high resolution spectra scales better with input size than bond-breaking.
CFM-ID takes on average 4.9 seconds per structure in the \textsc{NIST-20} test split, and scales quadratically $(R^2 = 0.78)$ with input size.
(We believe this is because larger molecules in \textsc{NIST-20} tend to be approximately path graphs -- e.g. long hydrocarbon chains -- with only quadratically many connected subgraphs.)
In comparison, running \textsc{GrAFF-MS} on the CPU takes 1.3 core-seconds per spectrum, and scales approximately linearly $(R^2 = 0.65)$.
This pays off at larger molecular weight: for molecules $> 500$Da, our model is $16\times$ faster on average.
Realistically, large-scale prediction will use the GPU: on a single GPU with batch size 512, predicting all of the \textsc{NIST-20} test spectra averages to 2.8ms per spectrum (mostly spent in preprocessing on the CPU).

We can additionally estimate the time required for each approach to generate a reference library from all 2,259,751 structures below 1000Da in ChEMBL v3.1.
Correcting for greater mean molecular weight in ChEMBL (405Da vs 292Da), this would take 2 hours running our PyTorch research code as-is on a single GPU.
The same library would take 4 \textit{days} to generate with 64 parallel instances of CFM-ID, which is written in optimized C++ code -- showing the importance of efficiently representing the output space.

\subsection{\textsc{GrAFF-MS} distinguishes very similar compounds and makes human-like mistakes}

\begin{figure}[ht]
\begin{center}
\centerline{\includegraphics[width=0.75\columnwidth]{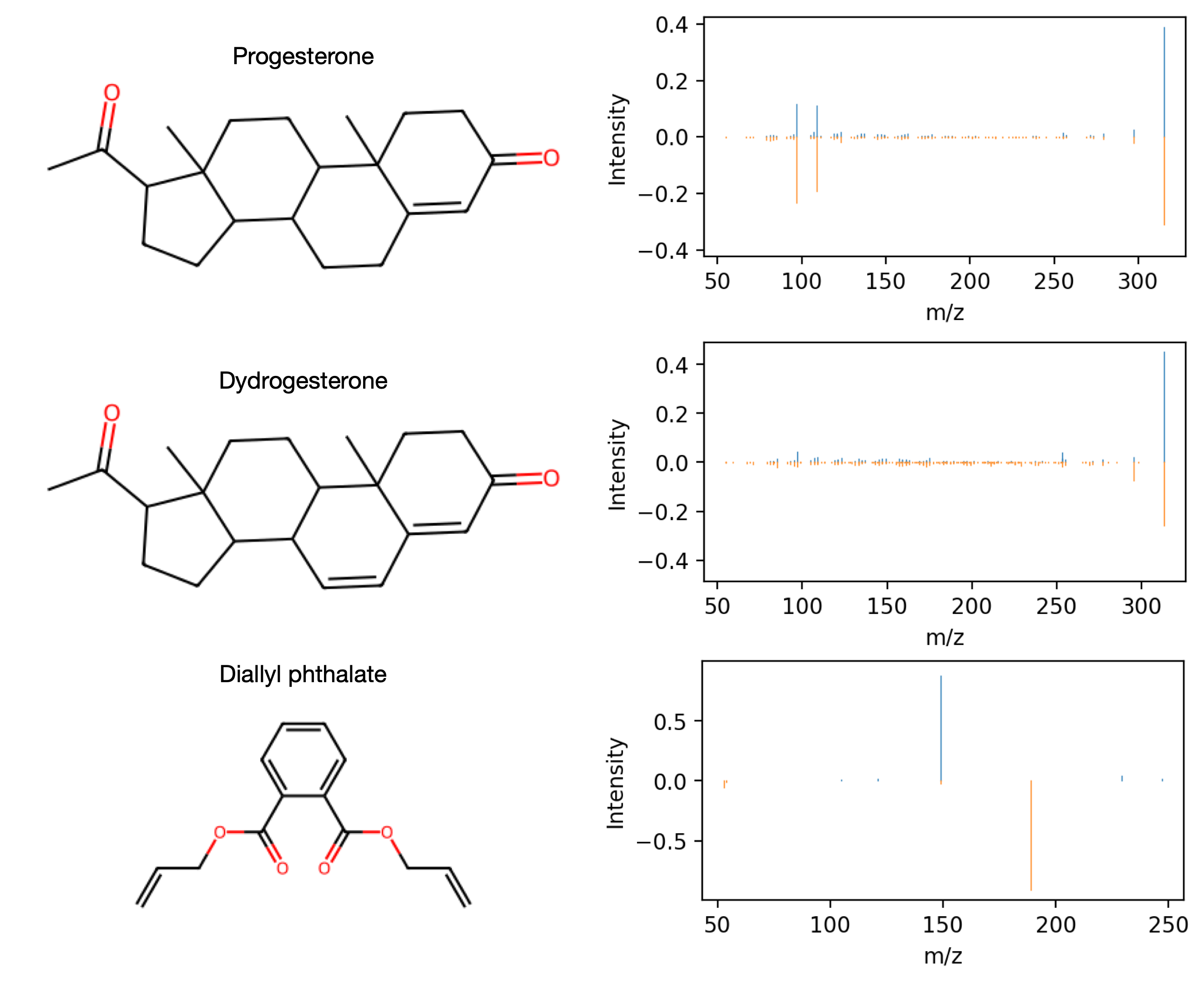}}
\caption{Three compounds from \textsc{CASMI-16}, with spectra predicted by our model (blue) against negated ground-truth (orange). Oxygens are shaded red by convention.}
\label{fig:spectra}
\end{center}
\end{figure}

In Figure \ref{fig:spectra} we show some particularly challenging examples of mass spectra.
The top and middle panels show two structurally similar compounds, differing only by the order of a carbon-carbon bond. 
Our approach correctly predicts distinct spectra for each ($C_{S\hat{S}} = 0.98$, top; $C_{S\hat{S}} = 0.95$, middle). 
The third molecule is an example where we fail to predict a realistic spectrum ($C_{S\hat{S}} = 0.03$), but in a manner in which a human expert would also fail.
This molecule is a member of the \textit{phthalate} class, which chemists recognize by a characteristic dominant peak at 149Da \cite{jeilani2011phthalate}. 
Our model predicts this same peak, correctly recognizing a phthalate: but in this case that peak is relatively minor, indicating atypical fragmentation chemistry.

\section{Discussion}

In this work we develop \textsc{GrAFF-MS}, a graph neural network for predicting high resolution mass spectra of small molecules.
Unlike previous approaches that force a tradeoff between $m/z$ resolution and a tractable learning problem, \textsc{GrAFF-MS} is both computationally efficient and capable of modelling the high-resolution $m/z$ information essential to mass spectrometry.
This is made possible by our discovery that mass spectra of small molecules can be closely approximated as distributions over a fixed vocabulary of molecular formulas, highlighting the value that domain-aware modelling can add to molecular machine learning. 
Particularly surprising was that we outperform CFM-ID, which trades model expressivity for an even stronger scientific prior that we expected would contribute to better generalization.
However, this prior incurs a heavy cost in time complexity, making it impractical to train CFM-ID on hundreds of thousands of spectra as we did.

Future directions include pretraining on large-scale property prediction tasks, revisiting sequential formula decoding, and incorporating additional scientific priors about fragmentation chemistry. Overall we anticipate this work will both accelerate scientific discovery and demonstrate mass spectrometry to be a compelling domain for continued machine learning research.

\bibliography{main}
\bibliographystyle{unsrt}

\clearpage

\appendix



\section{Fixed vocabulary selection}
\label{selection}

Algorithm \ref{alg:vocab} describes our procedure for selecting the product ions $\hat{\mathcal{P}}$ and neutral losses $\hat{\mathcal{L}}$.
We use the shorthand $\mathcal{F}^n_i = \mathcal{F}(P_n,m_i,\epsilon)$ to indicate the set of formulas computed by mass decomposition.
When mass decomposition yields more than one formula annotation for a peak, here we split the peak height uniformly among all annotations.

\begin{algorithm}[h]
   \caption{Fixed vocabulary selection}
   \label{alg:vocab}
\begin{algorithmic}
   \STATE {\bfseries Input:} training spectra and precursors $\{(S_n,P_n)\}_{n=1}^N$, vocabulary size $K$, tolerance $\epsilon$
   \STATE Initialize $F_f = 0, L_l = 0, \hat{\mathcal{P}} = \emptyset, \hat{\mathcal{L}} = \emptyset$
   \FOR{$n = 1 \dots N$}
   \FOR{$(m_i,y_i) \in S_n$}
   \FOR{$f \in \mathcal{F}^n_i$}
   \STATE $l = P_n - f$
   \STATE $F_f = F_f + y_i / \vert \mathcal{F}^n_i \vert$
   \STATE $L_l = L_l + y_i / \vert \mathcal{F}^n_i \vert$
   \ENDFOR
   \ENDFOR
   \ENDFOR
   \STATE Sort $F$ and $L$
   \WHILE{$\vert \hat{\mathcal{P}} \vert + \vert \hat{\mathcal{L}} \vert \le K$}
   \STATE $f = $ first element of $F$
   \STATE $l = $ first element of $L$
   \IF{$F_f > L_l$}
   \STATE Add $f$ to $\hat{\mathcal{P}}$
   \STATE Remove $f$ from $F$
   \ELSE{}
   \STATE Add $l$ to $\hat{\mathcal{L}}$
   \STATE Remove $l$ from $L$
   \ENDIF
   \ENDWHILE
\end{algorithmic}
\end{algorithm}

\section{Derivation of peak-marginal cross entropy}
\label{derivation}

We derive our loss function from physical first principles, making a number of minor modelling assumptions:
\begin{itemize}
\item The number of precursor ions accumulated by the instrument is Poisson with rate $\lambda$.
\item Each individual precursor ion is independently converted into fragment $f$ with probability $p_f$.
\item The instrument resolution parameter $\epsilon$ is sufficiently small that separate peaks do not overlap: there exists exactly one peak $i(f)$ for every $f: p_f > 0$ satisfying $\vert \langle\mu,f\rangle - m_i \vert \le \epsilon m_i$.
\end{itemize}
By the splitting property, the number of ions of each fragment are independently Poisson with rate $\lambda_f = \lambda p_f$.
By the merging property, the height of peak $i$ is also a Poisson r.v. $K_i$ with rate $\lambda_i = \sum_{f\in \mathcal{F}_i}\lambda_f$, where $\mathcal{F}_i$ denotes the set of fragments whose theoretical masses all fall within the measurement error $\epsilon m_i$ of peak $i$.

The log-likelihood for the peak height is (taking equality up to constants $C$ w.r.t. $p_f$):
\begin{align}
    & \log P(K_i = k_i) \\
    =&\ k_i \log \lambda_i - \lambda_i - \log k_i! \\
    =&\ k_i \log \left( \sum_{f \in \mathcal{F}_i} \lambda_f \right) - \left( \sum_{f \in \mathcal{F}_i} \lambda_f \right) + C \label{eq:merging} \\
    =&\ k_i \log \left( \sum_{f \in \mathcal{F}_i} \lambda p_f \right) - \left( \sum_{f \in \mathcal{F}_i} \lambda p_f \right)  \label{eq:splitting} \\
    =&\ k_i \log \lambda + k_i \log \left( \sum_{f \in \mathcal{F}_i} p_f \right) - \lambda \sum_{f \in \mathcal{F}_i} p_f \\
    =&\ C + k_i \log \left( \sum_{f \in \mathcal{F}_i} p_f \right) - \lambda \sum_{f \in \mathcal{F}_i} p_f
\end{align}
where (\ref{eq:merging}) uses merging, and (\ref{eq:splitting}) uses splitting. 
Because each fragment is assigned to exactly one peak (no overlap), the peak heights $\{K_i: i \in S\}$ are independent.
Let the total number of accumulated ions $K = \sum_{i\in S}k_i$ in spectrum $S$.
Defining $y_i = k_i / K$:
\begin{align}
    & \log P(\{K_i=k_i : i \in S\}) \\
    =& \sum_{i \in S} \log P(K_i = k_i) \\
    =&\ \sum_{i \in S} \left( k_i \log \left( \sum_{f \in \mathcal{F}_i} p_f \right) - \lambda \sum_{f \in \mathcal{F}_i} p_f \right) \\
    =&\ \sum_{i \in S} (K y_i) \log \left( \sum_{f \in \mathcal{F}_i} p_f \right) - \lambda \sum_{i \in S} \sum_{f \in \mathcal{F}_i} p_f \\
    =&\ K \sum_{i \in S} y_i \log \left( \sum_{f \in \mathcal{F}_i} p_f \right) - \lambda \cdot 1 \label{eq:unity} \\ 
    =&\ C \sum_{i \in S} y_i \log \left( \sum_{f \in \mathcal{F}_i} p_f \right) + C'
\end{align}
where (\ref{eq:unity}) again uses our assumption that every fragment is assigned to exactly one peak. 
Dropping the constants and negating the final term yields the peak-marginal cross-entropy loss for a single spectrum. $\qed$

\clearpage 

\section{Model hyperparameters}

We use a vocabulary of $K = 10000$ formulas.
We train an $L = 6$-layer encoder and $L' = 2$-layer decoder with $d_{enc} = 512$ and $d_{dec} = 1024$, resulting in $44.6$ million trainable parameters.
We use the $d_{eig} = 8$ lowest-frequency eigenvalues, truncating or padding with zeros.
All dropout is applied at rate $0.1$.
We use a batch size of $512$ and the Adam optimizer \cite{kingma2015adam} with learning rate $5\times 10^{-4}$.
We train for 100 epochs and use the model from the epoch with the lowest validation loss.
All models are trained using PyTorch Lightning with automatic mixed precision on 2 Tesla V100 GPUs.

\section{Mass spectral covariates}

\begin{table}[ht]
\caption{Mass spectral covariates used in our model.}
\label{tab:covs}
\begin{center}
\begin{small}
\begin{tabular}{lll}
\toprule
Feature & Range & Comment \\
\midrule
Normalized collision energy &
  $[0, 200]$ &
  \begin{tabular}[c]{@{}l@{}}Thermo Scientific PSB104, \\ ``Normalized Collision Energy Technology''\end{tabular} \\
  \\
Precursor type     & $[M+H]^+$, $[M-H]^-$ & Includes ionization mode $\&$ adduct composition \\
\\
Instrument model &
  \begin{tabular}[c]{@{}l@{}}Orbitrap Fusion Lumos, \\ Thermo Finnigan Elite Orbitrap, \\ Thermo Finnigan Velos Orbitrap\end{tabular} &
  Different limits of detection \\
  \\
Has isotopic peaks & False, True          & Proxy for width setting of precursor mass filter  \\
\bottomrule
\end{tabular}
\end{small}
\end{center}
\end{table}

\end{document}